# Orientation-Aware Planning for Parallel Task Execution of Omni-Directional Mobile Robot

Cheng Gong[1], *Student Member IEEE*, Zirui Li[1, 2], Xingyu Zhou[3], Jiachen Li[4], Junhui Zhou[3], Jianwei Gong[1], *Member IEEE*

*Abstract*— **Omni-directional mobile robot (OMR) systems have been very popular in academia and industry for their superb maneuverability and flexibility. Yet their potential has not been fully exploited, where the extra degree of freedom in OMR can potentially enable the robot to carry out extra tasks. For instance, gimbals or sensors on robots may suffer from a limited field of view or be constrained by the inherent mechanical design, which will require the chassis to be orientation-aware and respond in time. To solve this problem and further develop the OMR systems, in this paper, we categorize the tasks related to OMR chassis into orientation transition tasks and position transition tasks, where the two tasks can be carried out at the same time. By integrating the parallel task goals in a single planning problem, we proposed an orientation-aware planning architecture for OMR systems to execute the orientation transition and position transition in a unified and efficient way. A modified trajectory optimization method called orientation-aware timed-elastic-band (OATEB) is introduced to generate the trajectory that satisfies the requirements of both tasks. Experiments in both 2D simulated environments and real scenes are carried out. A four-wheeled OMR is deployed to conduct the real scene experiment and the results demonstrate that the proposed method is capable of simultaneously executing parallel tasks and is applicable to real-life scenarios.**

## I. INTRODUCTION

With feasibility being guaranteed, improving the efficiency in task execution has been one of the main objects for mobile robot systems. Researchers have been trying to improve the efficiency from different aspects. Generally, there are two aspects in improving the system efficiency for task execution, one is task planning which is considered more often for multi-robot systems, and the other is motion planning which is more often analyzed for single robots. Task planning methods aim to achieve optimal system efficiency at the task level, while motion planning tries to improve the efficiency in executing a specific task. In task planning, task allocation methods are widely researched and applied in multi-robot systems (MRS), which can increase the system efficiency by appropriately allocating the tasks to minimize the overall cost of execution. Task decomposition and task segmentation are also commonly used methods in task planning. The motivation of task decomposition methods is to divide complex tasks into multiple simpler parallel tasks that can be distributed to different robots. While task decomposition methods are mainly applied in multi-robot systems, task segmentation methods are usually applied for a single robot system, which aims to partition a difficult task into easier sequential tasks for a single robot to carry out. As for motion planning, many methods have been proposed to achieve better task performance through planning better motions for the executor, and a review on mobile robot motion planning methods can be found in [1].

Apart from the above two main aspects, some researchers have also paid attention to other aspects to further improve the system efficiency by parallelly executing task planning and motion planning[2-4]. Yet to the author's knowledge, few studies have paid attention to the parallel execution of multiple tasks in a single robot, especially when there are two or more executors available. One important reason may be that the chassis of mobile robots is difficult to be modeled in a unified way with other executors. However, for one special type of mobile robot, the omni-directional robot (OMR), which has three degrees of freedom, it is possible to execute extra tasks with the extra degree of freedom. Besides, the OMR can be modelled in a relatively easy and straightforward way, and can be seen as two executors in one unified model that has the potential to execute two parallel tasks parallelly. And due to its extreme flexibility in the narrow and complex indoor environment, OMRs have also been popular as research platforms and have been applied in various scenarios, such as service robot [5], medical support [6], on-site construction [7], human-robot collaboration [8], and robot competition [9, 10].

The unique omni-directional characteristic of OMR has also drawn many attentions in researching its dynamic model and motion planning schemes. In [11, 12], the Trajectory Linearization Control (TLC) methods were introduced based on dynamic model for motion control. After obtain the dynamic model with an arbitrary location of the center of mass, a smooth switching adaptive robust controller consisting of four parts was proposed in [13] for robust motion control. Besides, a Fuzzy-PI linear quadratic regulator controller was also proposed [14], and is applicable for embedded OMR systems. For path and motion planning, potential field methods were applied in [15, 16], where MPC controllers were combined for path tracking. After modelling the friction compensation, [17] proposed an extended state observer (ESO)

*This research was supported by the National Natural Science Foundation of China under grant number U19A2083. Corresponding author: Zirui Li and Jianwei Gong.

1 Cheng Gong, Zirui Li, Jianwei Gong are with the School of Mechanical Engineering, Beijing Institute of Technology, Beijing 100081, China. (e-mail: chenggong@bit.edu.cn; 3120195255@bit.edu.cn; gongjianwei@bit.edu.cn).

2  Zirui Li is also with the Department of Transport and Planning, Faculty of Civil Engineering and Geosciences, Delft University of Technology, Stevinweg 1, 2628 CN Delft, The Netherlands.

3 Xingyu Zhou, Junhui Zhou are with the School of Aerospace Engineering, Beijing Institute of Technology, Beijing 100081, China. (e-mail: zhouxingyu@bit.edu.cn; 3120200119@bit.edu.cn).

4 Jiachen Li is with  the Department of Mechanical Engineering, University of California, Berkeley, CA 94720, USA (e-mail: jiachen_li@berkeley.edu)

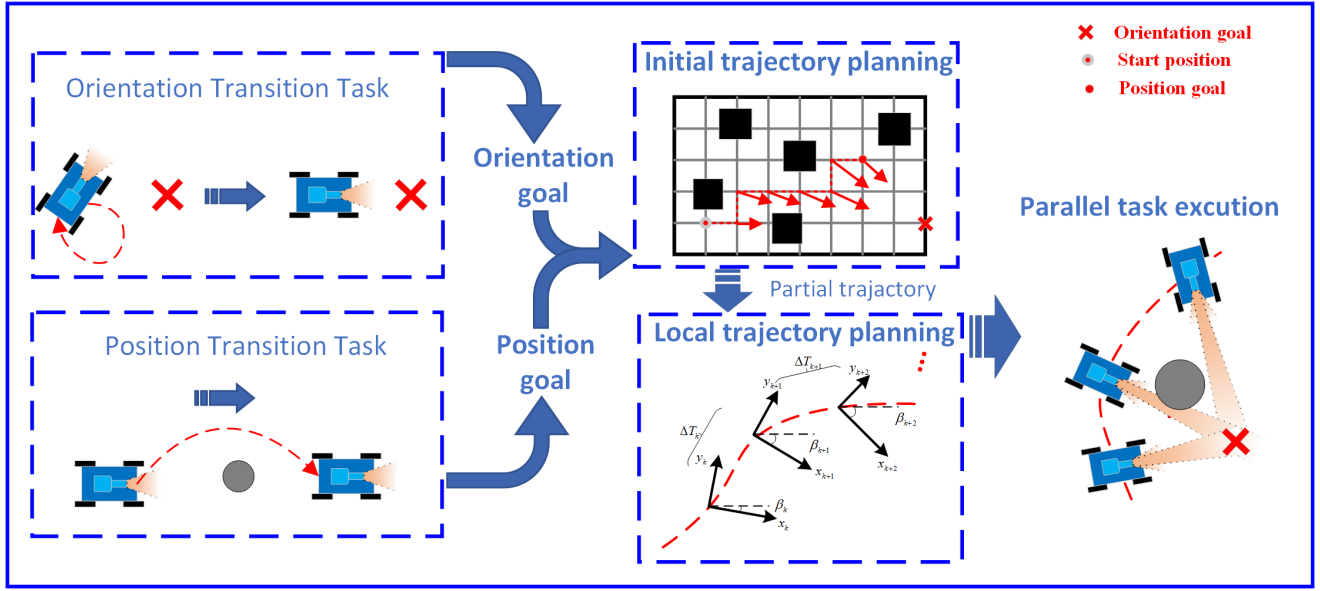

Fig.1 Illustration of parallel task execution procedure.

based SMC to control the OMR. Based on Non–linear Model Predictive Control, [18] introduced a fault tolerant control scheme for OMR. [19] also adopted MPC for trajectory tracking where delayed neural network (DNN) is applied in solving quadratic programming problem in optimization.

Although the above studies aim to exploit the flexibility and efficiency of OMR in application, their efforts remained at the motion planning related to position transition and did not consider orientation related tasks. In this paper, we propose a parallel task planning method which aims to further investigate the potential of OMR in multi-task execution by achieving the orientation transition task and position transition task in parallel. The proposed method is of great benefit in many circumstances, especially when the requirements of tasks are related to both position and orientation of the robot. For example, in OMR surveillance mission, where robots usually only have a limited field of view, and might lose sight of targets if the robot needs to transfer its position and the orientation transition is not well planned in movement. And this kind of problem can be solved with OMR by jointly considering the demand of the position transition task and the demand of the orientation transition task.

The illustration of the proposed method can be found in Fig.1, where the OMR tasks are decomposed into position transition tasks and orientation transition tasks. The goal of parallel tasks is extracted and integrated into a single motion planning problem. Bounded by the goals, the trajectory is firstly generated through the grid map approach in global scale and then finetuned using a time-optimal trajectory optimization method in the local scale. Based on Timed-Elastic-Band (TEB) [20], a modified online local trajectory optimization method called orientation-aware TEB (OATEB) is introduced. And to achieve both position transition demand and the orientation transition demand, kinematic constraint and obstacles are considered as well as two layers of orientation constraint in OATEB, with which the OMR can not only track the planned collision-free trajectory but also aim towards the expected orientation. The performance of the proposed method is validated in both the simulated environment and in the real scene. Experimental results demonstrated that the proposed method is adequate in executing position transition task and orientation transition task parallelly.

The paper is organized as follows: Section II formulates the basic idea of parallel task execution. Section III introduces the orientation-aware planning method. Section IV presents the experiments in both simulated environment and real scene, where experimental results are analyzed and discussed. Finally, section V summarizes the results and problems.

## II. PROBLEM FORMULATION

The omni-directional characteristic gives OMR extremely flexible maneuverability in plane ground, and makes it ideal for indoor complex tasks. To further exploit the flexibility and increase the overall task efficiency of OMR systems, we categorize OMR tasks into position transition and orientation transition tasks. As shown in Fig.1, OMR executes the orientation transition task by maintaining a specific orientation, for instance, surveilling a specific target. While in the position transition task, OMR aims to move to a target position without collision.

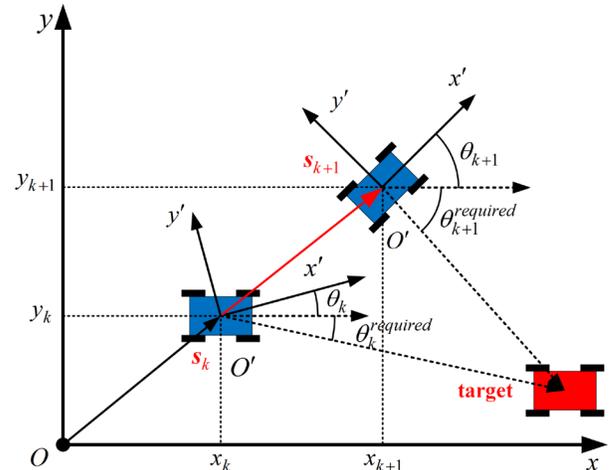

Fig. 2. Illustration of coordinate systems.

In application, these two tasks above usually are executed sequentially, since most car-like robots and differential-drive robots are non-holonomic constrained, where the position transition and orientation may contradict with each other. But for OMRs, which have three local degrees of freedom, these two tasks are possible to be executed in parallel, which can potentially increase the efficiency of the overall efficiency of the OMRs system.

In this paper, we aim to execute the position transition task and orientation transition task in parallel by integrating their goals in one motion planning problem. In other words, our goal is to control the OMR to move in a time-efficient and collision free manner while keep its orientation to the required direction. The description of the coordinate system in this paper is illustrated in Fig.2, where $O-xy$ represents the world coordinate and $O'-x'y'$ represents the robot coordinate. The robot pose at time step $k$ is described as $\mathbf{s}_k = [x_k, y_k, \theta_k] \in \mathbb{R}^3$, where $x_k, y_k$ denote the position under the world coordinate system, and $\theta_k$ describes the angle between the robot coordinate system and the world coordinate system. The conversion of the robot position between the robot coordinate system and the world coordinate system can be addressed as:

$$\begin{bmatrix} x' \\ y' \end{bmatrix} = Rot(\theta) \begin{bmatrix} x \\ y \end{bmatrix} = \begin{bmatrix} \cos\theta & \sin\theta \\ -\sin\theta & \cos\theta \end{bmatrix} \begin{bmatrix} x \\ y \end{bmatrix}, \quad (1)$$

where $Rot(\theta)$ represents the transfer matrix.

III. METHOD

In this section, we will firstly specify the goals of the position transition task and the orientation transition task in a unified representation in world coordinate system. Secondly, given the goals, an initial trajectory that satisfies both goals is generated using the grid map approach. Finally, an online trajectory planning method is modified and applied to optimize the local trajectory that satisfies the requirement of both the OMR kinemics and parallel task goals.

*A. Parallel goal specification*

To accomplish parallel task execution, goals of tasks need to be specified ahead. As described above, the position transition task and orientation transition task can be decomposed from OMR tasks as two isolated categories, where they do not affect each other in execution. For position transition, the goal specification can be relatively straightforward, which is to find an executable target position $\mathbf{p}_P = [x_P, y_P]$ that satisfies the requirement of the task. A grid map approach can be applied to search and locate the available target position and examine the executability of the target position. For position transition tasks with different goals or when goals need to be frequently updated, a behavior tree (BT) structure is applied in this paper to update the position transition goal.

Differing from the position transition task, the goal of orientation transition tasks is to maintain orientation of the OMR to a specific direction that is relative to both current robot position and the specific task requirement. To explicitly address the goal of orientation transition, in this paper we extract and present the goal of orientation transition with another position called orientation target position which is denoted as $\mathbf{p}_O = [x_O, y_O]$. By introducing this position, the goal of orientation transition tasks is transferred into a parallel configure which is to maintain the orientation towards the target position $\mathbf{p}_O$, for instance the geometric center of a surveillance target, while moving to the position target $\mathbf{p}_P$. And the execution of both tasks can be seen as a unified planning procedure. The real-time orientation requirement can be calculated using the current position $\mathbf{p}_k = [x_k, y_k] \in \mathbb{R}^2$ and orientation target position $\mathbf{p}_O$:

$$\theta_k^{required} = \arctan \frac{y_k - y_O}{x_k - x_O}, \quad (2)$$

*B. Initial trajectory planning*

After the position and orientation goals are obtained, the goals are applied in an initial trajectory planning in order to validate the executability of the given goal and provide an initial solution for the following time-optimal planning which will be detailed in the next part. To generate the initial trajectory, A* algorithm is applied in this paper to search the optimal path between the current position and the position goal in the grid map. The searched path of A* is a discrete set of positions, and can be described as:

$$\mathbf{P} = \{\mathbf{p}_k = [x_k, y_k] | k = 0, \ldots, f\} \quad (3)$$

where $\mathbf{p}_f$ is the goal of the initial path. The distance between $\mathbf{p}_f$ and $\mathbf{p}_P$ is controlled within a maximum distance $distance(\mathbf{p}_f, \mathbf{p}_P) \leq s^{tolerance}$, where $s^{tolerance}$ is set according to the grid size.

The orientation goal needs to be transferred to an OMR control config and integrated to the initial trajectory generation. And to cope with this, the discrete position of searched path can be extended to the robot pose which is a vector of three elements $\mathbf{s}_k = [x_k, y_k, \theta_k] \in \mathbb{R}^3$. Thus, the orientation is introduced in initial path generation. To accomplish the orientation goal, the expected heading angle is calculated as (2), and the initial trajectory set $\mathbf{S}^{initial}$ can be described as:

$$\mathbf{S}^{initial} = \left\{ \mathbf{s}_k^{ini} = \left[ x_k, y_k, \arctan \frac{y_k - y_O}{x_k - x_O} \right] \middle| k = 1, 2, \ldots, f \right\} \quad (4)$$

*C. Orientation-aware trajectory optimization*

After the initial trajectory is generated, we introduce the orientation-aware trajectory optimization method which takes the generated method trajectory as the initial solution and optimizes under both the position transition demand and orientation transition demand. Based on a time-optimal trajectory optimization method TEB, an orientation-aware modification called orientation-aware TEB (OATEB) is proposed. Two types of orientation constraint are applied in OATEB to achieve the purpose of fulfilling the orientation transition task.

As a local motion planner, OATEB has to preprocess the generated initial path before using it as an initial solution for the trajectory optimization. To reduce the calculation complexity, only a limited near-by area will be considered and modeled for trajectory optimization. And the initial path is first sectioned into small segments with designated length $L$, where only the first segment is considered, which can be described as $\mathbf{S}_1^{initial} = [\mathbf{s}_0^{ini}, \mathbf{s}_1^{ini}, ..., \mathbf{s}_i^{ini}, ..., \mathbf{s}_{L-1}^{ini}]$. The discrete trajectory $\mathbf{S}_1^{initial}$ with $L$ pose is then used as an initial solution for the local trajectory optimization, where the discrete pose of initial path will be the edge for the local trajectory generation and bound the optimized trajectory to meet the requirement of orientation transition task.

In TEB method, the presentation of planned trajectory is extended with time interval $\Delta T_k \in R^+$, $k = 1, 2, ..., n-1$ and is formed as a vector $\mathbf{b}$:

$$b = [s_1, \Delta T_1, s_2, \Delta T_2, s_3, ..., \Delta T_{n-1}, s_n]^T. \quad (5)$$

The basic idea of OATEB is consistent with TEB, which is to generate a time-optimal trajectory. And with time as the optimization goal, the problem of trajectory planning can be converted to a nonlinear optimization problem:

$$V^*(\mathbf{b}) = \min_{\mathbf{b}} \sum_{k=1}^{n-1} \Delta T_k^2$$

$$\begin{aligned}
s.t. \quad & \mathbf{s}_1 = \mathbf{s}_0, \mathbf{s}_n = \mathbf{s}_f \\
& \mathbf{v}_k(\mathbf{s}_k, \mathbf{s}_{k+1}) \geq 0 && k = 1, 2, \cdots, n-1 \\
& \mathbf{a}_k(\mathbf{s}_k, \mathbf{s}_{k+1}, \mathbf{s}_{k+2}) \geq 0 && k = 1, 2, \cdots, n-2 \quad (6) \\
& \boldsymbol{\omega}_k(\mathbf{s}_k, \mathbf{s}_{k+1}) \geq 0 && k = 1, 2, \cdots, n-1 \\
& \boldsymbol{\alpha}_k(\mathbf{s}_k, \mathbf{s}_{k+1}, \mathbf{s}_{k+2}) \geq 0 && k = 1, 2, \cdots, n-2 \\
& \mathbf{o}_k(\mathbf{s}_k) \geq 0 && k = 1, 2, \cdots, n \\
& \boldsymbol{\theta}_k(\mathbf{s}_k) \geq 0 && k = 1, 2, \cdots, n
\end{aligned}$$

where several constraints should be met in order to ensure that the trajectory is feasible and satisfies the requirement of the parallel tasks. First of all, the processed initial trajectory is applied as an initial solution in this problem, and the start pose and end pose of the initial trajectory are also set as the edge constraint for the local trajectory:

$$\mathbf{s}_1 = \mathbf{s}_0^{ini}, \quad \mathbf{s}_n = \mathbf{s}_{L-1}^{ini}, \quad (7)$$

where $\mathbf{s}_0^{ini}$ describes the initial state and $\mathbf{s}_{L-1}^{ini}$ describes the end state, and are extracted from the initial solution $\mathbf{S}_1^{initial}$. Since the speed tends to be maximized in time-optimal planning, the speed constraint $\mathbf{v}_k(\mathbf{s}_k, \mathbf{s}_{k+1}) \geq 0$ is also applied in OATEB to ensure the feasibility of the generated trajectory, which can be described as:

$$\mathbf{v}'_k = Rot(\theta_k)\mathbf{v}_k \leq \mathbf{v}'_{max}, \quad (8)$$

where $\mathbf{v}'_k = [v'_{x,k}, v'_{y,k}]$ is the robot velocity under the robot coordinate system and $\mathbf{v}'_{max} = [v'_{x\,max}, v'_{y\,max}]$ represents the maximum linear speed of the OMR along the direction of $x'$ axis and $y'$ axis, and $\mathbf{v}_k = [v_{x,k}, v_{y,k}]$ is the robot velocity under the coordinate system in each time step and can be calculated as:

$$\mathbf{v}_k = \Delta T_k^{-1}[x_{k+1} - x_k, y_{k+1} - y_k], \quad (9)$$

To ensure the output torque is executable, the acceleration constraint $\boldsymbol{\alpha}_k(\mathbf{s}_k, \mathbf{s}_{k+1}, \mathbf{s}_{k+2}) \geq 0$ is applied and can be described as:

$$\mathbf{a}'_k = Rot(\theta_k)\mathbf{a}_k \leq \mathbf{a}'_{max} \quad (10)$$

where $\mathbf{a}'_k = [a'_{x,k}, a'_{y,k}]$ is the robot linear acceleration and $\mathbf{a}'_{max} = [a_{x\,max}, a_{y\,max}]$ is the maximum acceleration along the $x'$ and $y'$ direction in the robot coordinate system. $\mathbf{a}_k = [a_{x,k}, a_{y,k}]$ is the acceleration under the world coordinate system and can be calculated as:

$$\mathbf{a}_k = \frac{2(\mathbf{v}_{k+1} - \mathbf{v}_k)}{\Delta T_k + \Delta T_{k+1}} \quad (11)$$

Similarly, there is angular speed constraint $\omega_k(\mathbf{s}_k, \mathbf{s}_{k+1}) \geq 0$, and angular acceleration constraints $\alpha_k(\mathbf{s}_k, \mathbf{s}_{k+1}, \mathbf{s}_{k+2}) \geq 0$:

$$\omega_k = \frac{\theta_{k+1} - \theta_k}{\Delta T_k} \leq \omega_{max}, \quad (12)$$

$$\alpha_k = \frac{\omega_{k+1} - \omega_k}{\Delta T_k} \leq \alpha_{max}, \quad (13)$$

where $\omega_{max}$ and $\alpha_{max}$ are the maximum angular speed and maximum angular acceleration of OMR. $\omega_k$ and $\alpha_k$ represent the angular speed and acceleration at pose $\mathbf{s}_k$.

To fulfill the position transition task, collision should be avoided, and obstacles constraint $\mathbf{o}(\mathbf{s}_k) \geq 0$ is introduced to ensure the planned trajectory is collision-free, where the pose in each time step is bounded by:

$$\rho(\mathbf{s}_k, \mathcal{O}^i) \geq R_{ROI}, \quad i = 1, 2, ... \quad (14)$$

where $\rho(\mathbf{s}_k, \mathcal{O}^i)$ is minimum Euclidean distance between the obstacle $\mathcal{O}^i$ and the robot pose $\mathbf{s}_k$.

Beside position transition task, the planned trajectory should also meet the requirement of the orientation transition task. To make sure the continuous execution of orientation transition, besides the orientation boundary from the initial trajectory poses, we applied an orientation constraint $\theta_k(\mathbf{s}_k) \geq 0$ to each pose in the optimization of trajectory, where the orientation of each pose should be satisfied:

$$\left|\theta_k - \theta_k^{required}\right| \leq \Delta\theta_{max} \quad (15)$$

where $\theta_k^{required}$ is the required orientation, and $\Delta\theta_{max}$ represents the maximum orientation error that can be accepted in practice. After adding above constraints as a penalty function to the optimization goal, the penalty function of the inequality constraint can be marked as:

$$\chi(\mathbf{c}, \sigma_c) = \sigma_c \|\min\{0, \mathbf{c}\}\|_2^2 \quad (16)$$

where $\mathbf{c}$ represents the constraint and $\sigma_c$ is the weighted factor of constraint $\mathbf{c}$ in optimization. Considering the penalty

function, the trajectory planning problem with approximate least squares can be described as:

$$b^* = \arg \min_{S \setminus \{s_1, s_2\}} \tilde{V}(b) \quad (17)$$

where $b = S \setminus \{s_1, s_2\}$ is the trajectory description parameter, and the expression of $\tilde{V}(b)$ is as follows:

$$\tilde{V}(b) = \sum_{k=1}^{n-1} [\Delta T_k^2 + \chi(v_k, \sigma_v) + \chi(a_k, \sigma_a) + \chi(\omega_k, \sigma_\omega) \cdots \\ + \chi(\alpha_k, \sigma_\alpha) + \chi(o_k, \sigma_o) + \chi(\theta_k, \sigma_\theta)] \quad (18)$$

The optimization problem can be solved using graph optimization toolkit g2o [21]. The result obtained by the g2o toolkit solution is the timed trajectory presentation $b$, which can then be transferred into robot motion in the world coordinate system and then converted to the velocities of the robot $v_x', v_y', \omega$.

## IV. EXPERIMENTS AND DISCUSSION

In this section, the proposed parallel execution planning method is evaluated through experiments and results are presented and analyzed. Both simulated environment and real scene environment are adopted and configured in experiments. The experiments are firstly carried out in the simulated environment to examine the performance and executability of the proposed method. Then, experiments are conducted in the real scenario to further illustrate its applicability on real-world application. For each experimental environment, experimental settings are introduced firstly, which include the experimental platform, the parallel task setting, parameters of the algorithm, etc. Then, experimental results are presented and analyzed, where the performance of both the position transition task and the orientation transition task will be presented and validated. Finally, the overall task execution ability of the proposed method to carry out the parallel task execution is validated and analyzed.

### A. Experiment in the simulated environment

#### 1) Experimental settings

The simulated environment is built based on *stage_ros*, which is a 2D simulation platform in ROS. In the simulation environment, the kinematic model of the robot, and the collision checking between the robot and the environment obstacles are also considered. The experimental environment is illustrated in Fig.3, which is set up as a rectangular area that is 8m long and 5m wide and with several rectangular obstacles in it. Each grid in Fig.3 represents 1 m, obstacles and boundary of the ground are marked in black, and the robot size is considered for judging collision with obstacles or border. The length and width of the robot is 0.6 m and 0.45 m respectively, and the collision size is inflated by 0.4 m to balance the error in control and localization.

The parallel task setting in this experiment is also shown in Fig.3, where the position transition goal is for the robot is to move from start to end positions and the orientation transition goal is to keep orientation towards the target enemy in order to monitor or attack the target. Since the robot visual sensor has only a limited field of 75 degrees, the robot should keep its expected orientation towards the enemy target within an acceptable error of around 30 degrees. To demonstrate the performance of the proposed method in parallel task execution, a planning method that uses a simple TEB and does not consider parallel task execution is also applied as a comparative method in the same scenario.

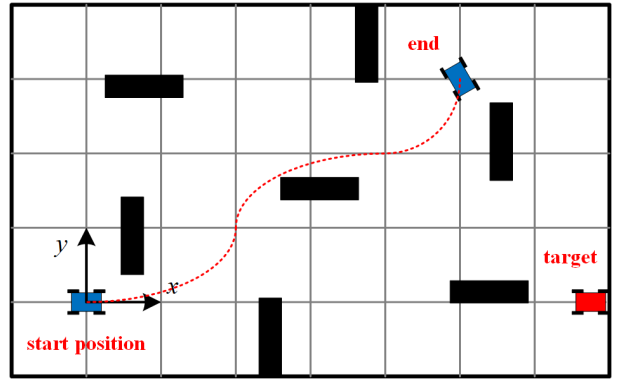

Fig. 3. Simulation environment and parallel task settings.

### B. Experimental result in simulated environment

Parameters of OATEB algorithm are as follows: $\sigma_v = 1$, $\sigma_a = 3$, $\sigma_\omega = 1$, $\sigma_\alpha = 3$, $\sigma_o = 50$ and $\sigma_\theta = 1$. The max tolerated orientation error in constraint is set to $\theta_{max} = 15$ degrees. Fig.4 illustrates the planning trajectory generated by the proposed method that is obtained in Rviz, where dark zones represent obstacles, and the light blue zones are the inflated zones of obstacles. The origin of the arrow represents the position of the robot, and the direction of the arrow represents the orientation of the robot at the current position. It can be observed from Fig.4 that the robot's trajectory meets the obstacle avoidance requirements. In addition, the orientation of the robot in trajectory basically meets the demand of orientation transition task.

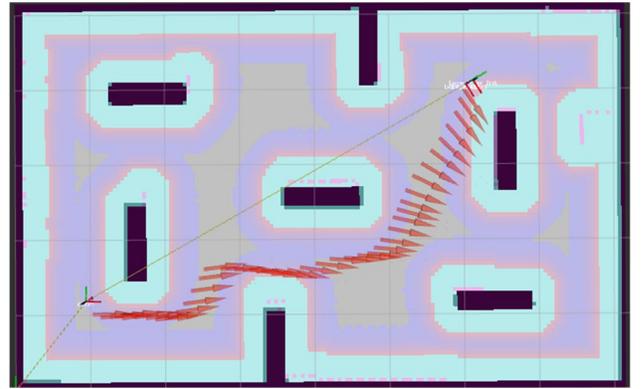

Fig. 4. The visualization of generated sparse trajectory in Rviz.

The executed orientation and the orientation error are shown in Fig.5, where $\theta^{real}$ is the real orientation of the robot and $\theta^{required}$ represents the required and the resulting trajectory orientation in the task. $\Delta \theta = |\theta^{real} - \theta^{required}|$ represents the orientation error, and $\theta^{max}$ is the maximum error allowed in the orientation constraint. It can be observed that the trajectory generated by proposed method meets the requirements of orientation constraints where the orientation error keeps a lower level than the constraint of 15 degrees.

The comparison between the trajectory and orientation of the proposed method and compared method is illustrated in Fig.6, where the proposed method is marked as OATEB, and the compared method is marked as TEB for better presentation. It can be observed that the trajectory of OATEB and TEB are similar but the OATEB can control the orientation error within a minor range. And it can be found that the trajectory obtained from the solution satisfies the beginning and end constraints and is collision free. However, since the TEB algorithm does not consider the orientation constraint, the robot is oriented in an arbitrary direction during the task, and only turns to the target orientation when close to the end position. Compared to the result of OATEB, the robot lost sight of the target when moving to the end position.

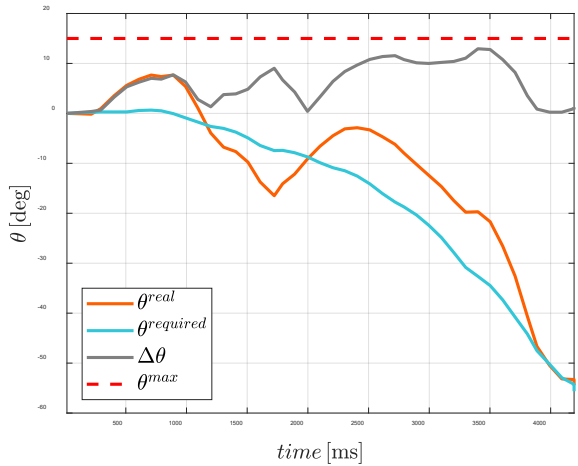

Fig. 5. Orientation constraint test of OATEB

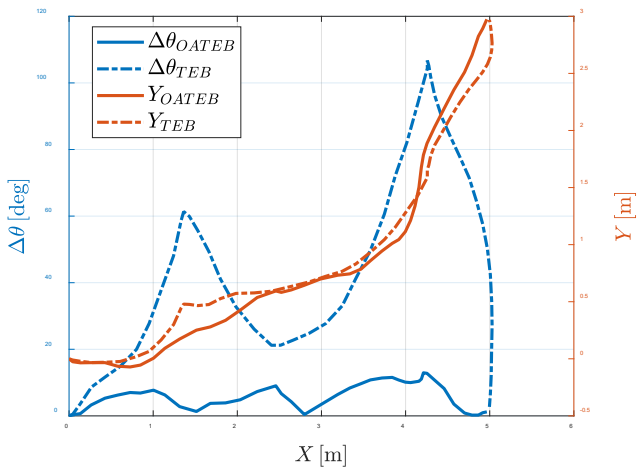

Fig. 6. Comparisons between planned trajectories (orange) and orientation error (blue) of OATEB (solid) and TEB (dash).

It is worth noting that, compared to OATEB, the trajectory planned by TEB takes more time to execute position transition task in the above simulation. One possible explanation is that, although the waypoints obtained from global planner are the same for both methods, the TEB algorithm only takes an orientation constraint at the end position. Thus, using TEB the robot needs to rotate according to the end orientation constraint when it approaches the end position. Instead, the OATEB algorithm can adjusts the orientation in real time and tracks the target along the movement, which segments one severe rotation action into minor actions during the movement process, resulting in less action time.

In OATEB, different orientation constraint $\theta^{max}$ has different influence in performance of OATEB, and results of $\theta^{max} = 5, 10, 15, 20$ degrees are shown in Fig.7. As the value of $\theta^{max}$ increases, the orientation error of the planned trajectory also increases accordingly. And theoretically, the OATEB is more likely to find a more time-efficient path when the orientation constraint is looser, where the result of $\theta^{max} = 20$ degree achieves the best time efficiency. In practical use, the constraint can be modified according to the specific orientation task request to balance the performance in orientation error and time-efficiency.

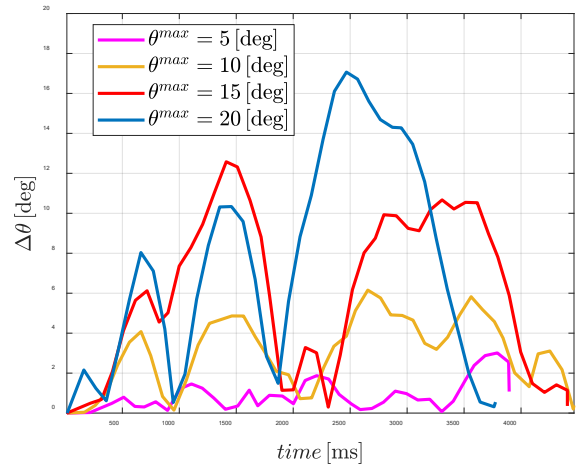

Fig. 7. OATEB performance under different orientation constraint

### C. Experimental results in real scene

#### 1) Experimental setting

The experimental environment for real scene evaluation is presented in Fig.8. The experimental scene is an indoor area that is similar to the settings of the simulated experiment. The experimental platform is based on *DJI RoboMaster 2019 AI Robot Platform*, which has a four-wheeled omni-direction chassis and a gimbal that can fire plastic rounds. A LiDAR is fixed to the chassis in the front, and a camera is attached to the axis of gimbal. The UWB system is also deployed for robot localization. The robot software architecture is built based on RoboRTS, which is an open source software stack developed by RoboMaster. An Intel NUC8i5BEK mini PC equipped with an Intel® Core™ i5-8259U Processor is adopted to provide computing platform for the software.

Different from the experiment in simulated environment, the goal of orientation transition task is set to the center of the experimental area. And the position transition task is to transit between the four corners of the experimental site, where the position transition goal is given by a behavior tree node that gives fixed goals in order. The localization result and camera image are also collected for the parallel task execution performance evaluation.

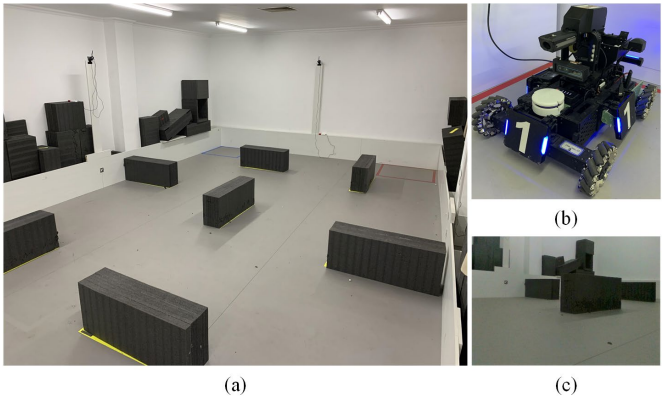

Fig. 8. Experimental environment and platform in real scene, (a) the experiment site, (b) the *DJI Robomaster* robot platform equipped with camera and LIDAR, (c) the camera vision.

*2) Experimental results*

The trajectory and pose of the robot in parallel task execution are illustrated in Fig.9, where the robot pose is selected evenly from the robot localization result in time sequence. As the figure shows, the green arrows that represent the orientation of the robot are pointed to the target position of orientation transition goal. And the trajectory of the robot did not collide with obstacles on site while and the orientation of the planned trajectory basically meets the requirement of orientation transition task, which is to keep the orientation of the robot to the center of the experiment site.

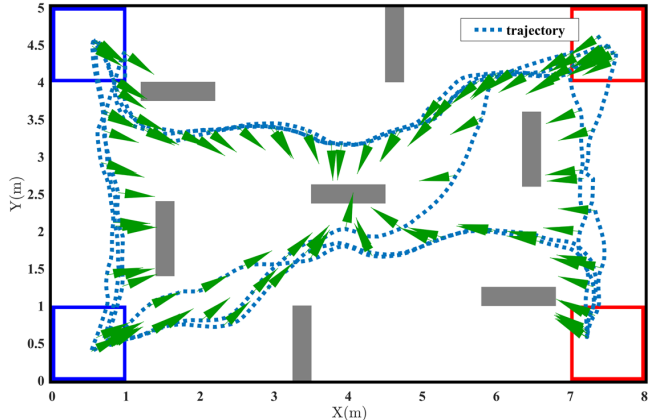

Fig. 9. Trajectory and pose of the robot in real scenario experiment. The pose amount is reduced for better visual representation.

The detailed experiment results are shown in Fig.10, where the orientation, velocity and acceleration information are presented. The executed orientation $\theta^{real}$ and the relative orientation that is required by the orientation transition task $\theta^{require}$ is shown in Fig.10 (a). The difference between the executed orientation and required orientation is also presented in this figure, which can be seen as the error in orientation transition task and is shown as the purple solid area in the figure. Fig.10 (a) also shows that the robot can execute the orientation transition task with a relatively low level of error in most cases. The average error of orientation transition task in this experiment is $\Delta\theta = 0.0875$ rad and is about 5 degrees, which is sufficient for application like surveillance.

In the real scene experiment, it can be observed that the error can be relatively high in some cases. For instance, the error $\Delta\theta$ at 5.6s is 0.8049 rad which is about 50 degrees. This happened when the robot is very close to the orientation target position, and it can be referred to Fig.10 (b) that the angular velocity at the time is relatively high. And Fig.10 (b)-(c) shows that high linear velocity and high acceleration have no strong impact in increasing the orientation transition error. Thus, a constraint in lower angular velocity may improve the orientation transition performance but may undermine the efficiency of position transition performance at the same time.

V. CONCLUSION

In this work, our goal is trying to increase system efficiency by parallelly executing both position transition and orientation transition tasks. An orientation-aware planning architecture for parallel execution is proposed in this paper, where the parallel tasks are integrated into one motion planning problem and a modified trajectory optimization method OATEB is introduced. Both experiments in the simulated environment and the real scene are carried out. Experimental results prove that the proposed algorithm is capable of generating feasible trajectories that enable the robot to maintain the expected orientation while transiting its position in harmony. In the real scene, the average orientation error of the method can be control around 5 degrees and can be further decreased with lower angular velocity, which is enough for orientation transition tasks that do not require high precision. Future works will focus on applying the orientation-aware planning architecture in urban interactive scenarios by considering the prediction of traffic participants [22-25].

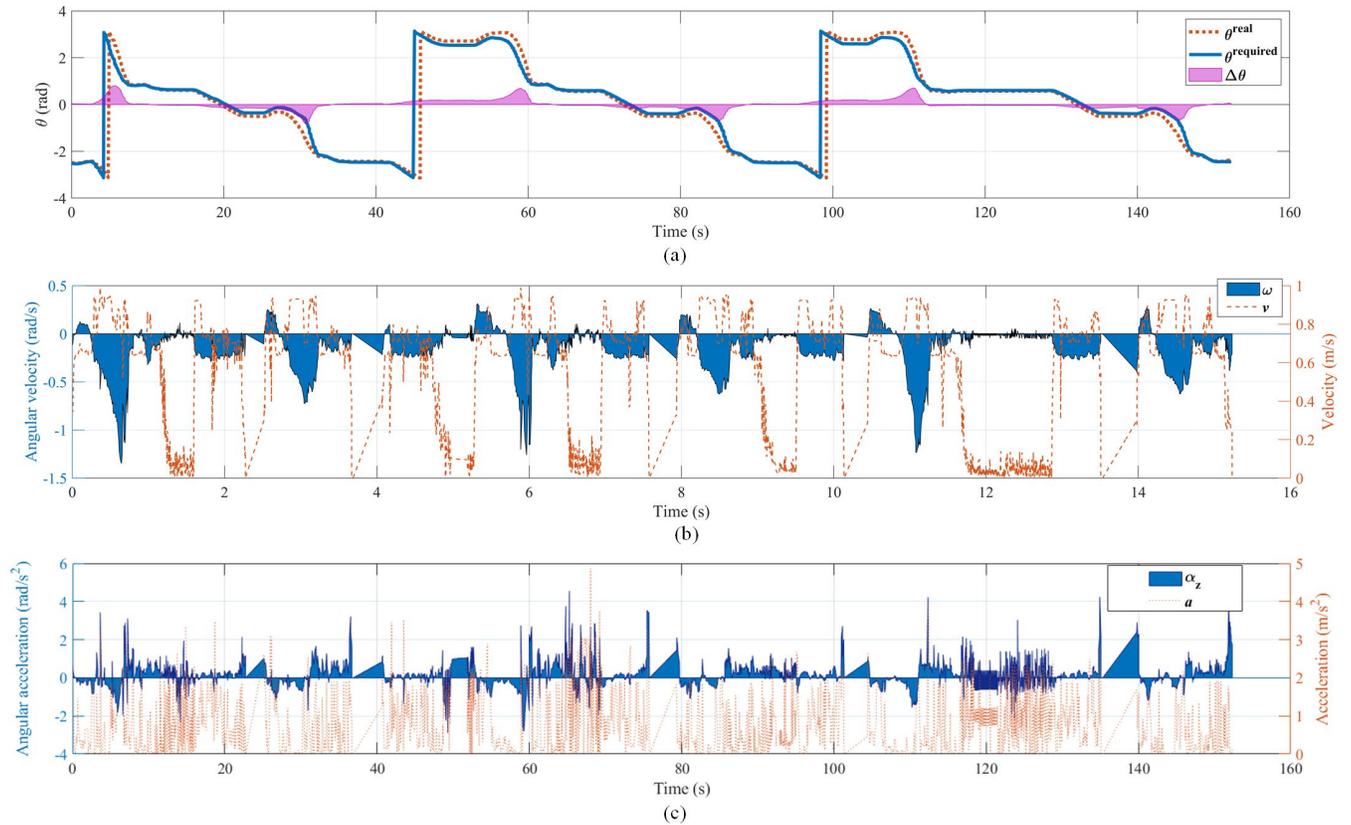

Fig. 10. Experimental result in real scenario experiment, (a) shows the real executed robot orientation $\theta^{real}$ and the required orientation $\theta^{required}$ and their difference $\Delta\theta$, (b) shows the executed linear velocity $v$ and angular velocity $\omega$, (c) shows the executed linear acceleration $a$ and angular acceleration $\alpha_z$.